\documentclass[final]{cvpr}

\usepackage{times}
\usepackage{epsfig}
\usepackage{graphicx}
\usepackage{amsmath}
\usepackage{amssymb}
\usepackage{multirow}
\usepackage{booktabs}
\usepackage[percent]{overpic}

\usepackage{array}
\usepackage{color, colortbl}
\usepackage{pifont}
\newcommand{\cmark}{\ding{51}}
\newcommand{\xmark}{\ding{55}}
\newcolumntype{R}{@{\extracolsep{0.2cm}}r@{\extracolsep{0pt}}}

\newcommand{\gt}{ground-truth}
\newcolumntype{;}{!{\vrule width 2pt}}

\definecolor{LightYellow}{rgb}{1,1,0.7}

\usepackage[pagebackref=true,breaklinks=true,colorlinks,bookmarks=false]{hyperref}

\begin{document}

\title{Learning optical flow from still images}

\author{Filippo Aleotti\thanks{Joint first authorship.} \hspace*{1cm} Matteo Poggi$^*$ \hspace*{1cm} Stefano Mattoccia\\
Department of Computer Science and Engineering (DISI)\\
University of Bologna, Italy\\
{\tt\small \{filippo.aleotti2, m.poggi, stefano.mattoccia \}@unibo.it}
}

\maketitle

\begin{abstract}
This paper deals with the scarcity of data for training optical flow networks, highlighting the limitations of existing sources such as labeled synthetic datasets or unlabeled real videos.
Specifically, we introduce a framework to generate accurate ground-truth optical flow annotations quickly and in large amounts from any readily available single real picture. Given an image, we use an off-the-shelf monocular depth estimation network to build a plausible point cloud for the observed scene. Then, we virtually move the camera in the reconstructed environment with known motion vectors and rotation angles, allowing us to synthesize both a novel view and the corresponding optical flow field connecting each pixel in the input image to the one in the new frame.
When trained with our data, state-of-the-art optical flow networks achieve superior generalization to unseen real data compared to the same models trained either on annotated synthetic datasets or unlabeled videos, and better specialization if combined with synthetic images.
\end{abstract}

\section{Introduction}

The problem of estimating per-pixel motion between video frames, also known as \textit{optical flow} \cite{sun2010secrets}, has a long history in computer vision and remains far from being solved. On top of it, several higher-level tasks such as tracking, action recognition and more are typically performed.
Among the main challenges for optical flow systems, there are occlusions, motion blur and lack of texture.

Deep learning has played a crucial role in the latest years of research on this topic, at first to learn a data term \cite{bai2016exploiting,xu2017accurate} and then to directly infer the dense optical flow field in end-to-end manner \cite{dosovitskiy2015flownet,ilg2017flownet2,sun2018pwc,sun2019models,hui18liteflownet,hui20liteflownet2,hui20liteflownet3,teed2020raft}, currently representing the state-of-the-art in this field. This achievement has been made possible by the availability of extensive training data labeled with \gt{} optical flow fields, most of them obtained through computer graphics \cite{butler2012sintel,dosovitskiy2015flownet,ilg2017flownet2}.
\begin{figure}
    \centering
    \renewcommand{\tabcolsep}{1pt}
    \begin{tabular}{cccc}
        \includegraphics[width=0.11\textwidth]{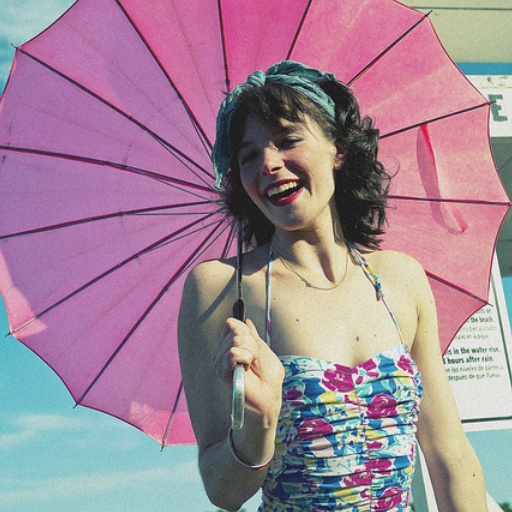} &
        \includegraphics[width=0.11\textwidth]{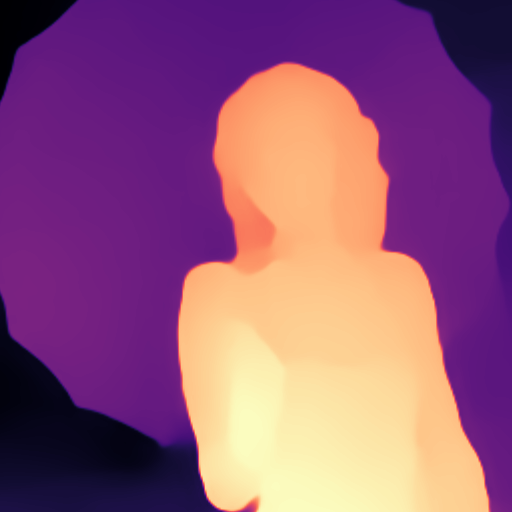} &
        \includegraphics[width=0.11\textwidth]{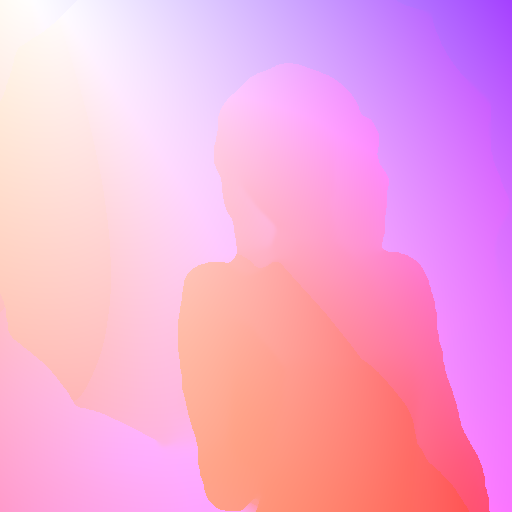} &
        \includegraphics[width=0.11\textwidth]{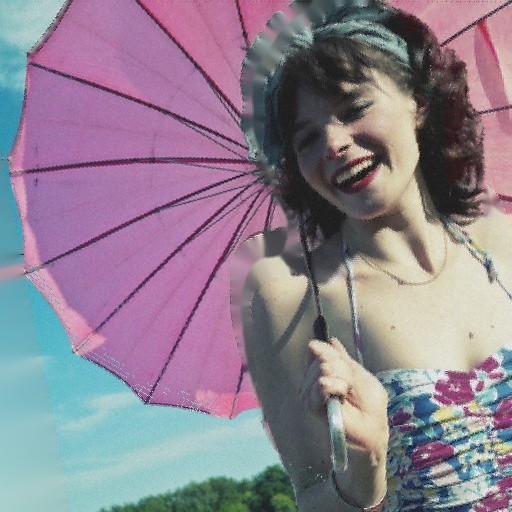} \\
        \includegraphics[width=0.11\textwidth]{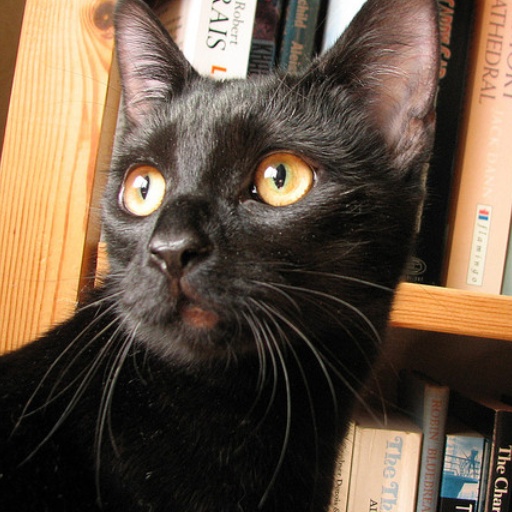} &
        \includegraphics[width=0.11\textwidth]{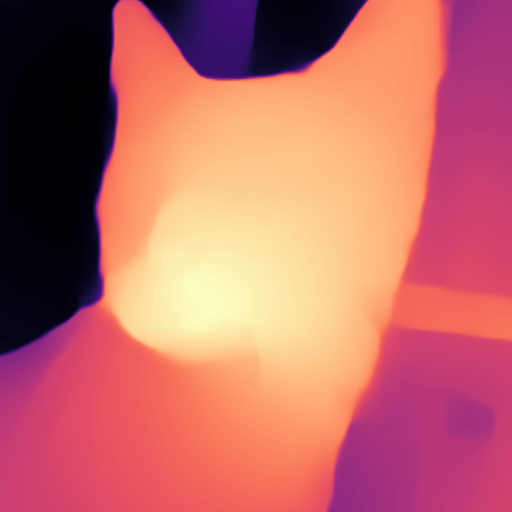} &
        \includegraphics[width=0.11\textwidth]{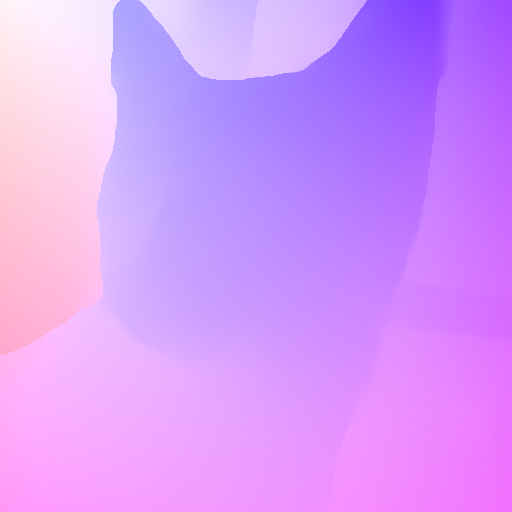} &
        \includegraphics[width=0.11\textwidth]{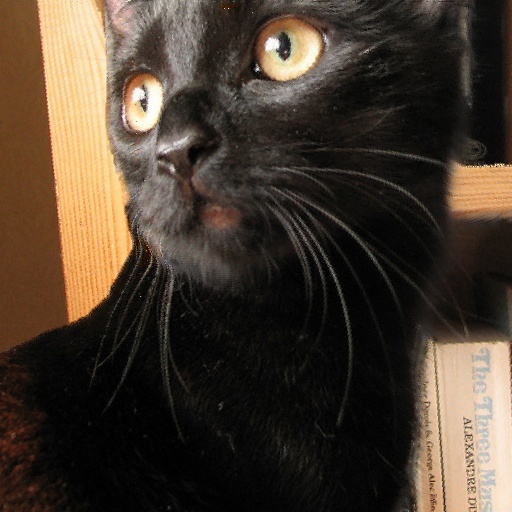} \\
        a) & b) & c) & d) \\
    \end{tabular}
    \caption{\textbf{\textit{Depthstillation} from still images.} From left to right: a) single input image, b) estimated depth map, c) optical flow field consequence of virtual camera motion, d) virtual view. We show b) as inverse depth to improve visualization.}
    \label{fig:banner}
\end{figure}
Unfortunately, these large datasets alone are not enough to train a neural network for its deployment in real environments, because of the well-known \textit{domain shift} occurring when moving from synthetic images to real ones. A notable example is represented by the KITTI optical flow benchmarks \cite{geiger2012kitti,menze2015kitti}, over which deep networks that have been trained only on synthetic data perform poorly, as witnessed by recent works \cite{ilg2017flownet2,sun2018pwc,teed2020raft}.
This problem is known in literature and has been faced for other tasks such as semantic segmentation \cite{hoffman2018cycada,murez2018image,ramirez2019learning,toldo2020unsupervised} or stereo depth estimation \cite{tonioni2017unsupervised,Tonioni_2019_CVPR,zhang2019domain,watson2020stereo}. To fully restore a level of accuracy comparable to the one achieved on synthetic data, fine-tuning on imagery similar to the testing domain is usually required. Anyway, obtaining \gt{} optical flow labels for real images is particularly challenging because there exists virtually no sensor capable of acquiring \gt{} correspondences between points in challenging real-world scenes \cite{Menze2018JPRS}. A viable strategy consists into passing through depth sensors (\eg, LiDARs), indeed optical flow fields can be obtained by projecting the 3D points from a given frame into the next frame \cite{geiger2012kitti}, although it cannot take into account independently moving objects, for which manual post-processing or annotation remains necessary \cite{menze2015kitti,Menze2018JPRS}. The literature is rich of self-supervised strategies \cite{meister2018unflow,liu2019selflow,liu2020learning,jonschkowski2020uflow} from unlabeled videos to soften this constraint, but they mostly excel when deployed on data similar to those observed for training, a scenario unlikely to occur in most real applications.

Given both the aforementioned domain shift issue and the lack of real imagery annotated for optical flow, we propose an alternative scheme to distill proxy labels from real images for effective training of optical flow estimation networks. Following the observation that depth is required to obtain dense matching across views through reprojection \cite{geiger2012kitti,menze2015kitti,Menze2018JPRS}, we use a monocular depth estimation network to revert the annotation process: given a single image and its estimated depth, we suppose a \textit{virtual} motion of the camera to compute a dense optical flow field and, consequently, synthesize a new \textit{virtual} image accordingly.
For instance, in Figure \ref{fig:banner} from a) pictures of a person and a cat, we estimate b) monocular depth and generate c) a flow field used to synthesize d) a novel view. We dub this process \textit{Depthstillation}, and any single image is eligible for producing optical flow annotations through it.

Experiments carried out on synthetic (Sintel) and real (KITTI 2012 and 2015) datasets support our main claims:
\begin{itemize}
    \item We show that it is possible to train an optical flow network on a collection of unrelated images, \eg single pictures readily available online

    \item Using real images through our technique allows us to train networks that better transfer to real data than their counterparts trained on synthetic images, while fine-tuning these latter on \textit{dephtstilled} frames and then on real data improves specialization
    
    \item Networks trained on our dephtstilled frames and flow labels better transfer to new real datasets than state-of-the-art self-supervised strategies using real videos \cite{jonschkowski2020uflow}

\end{itemize}

\section{Related Work}

In this section, we review the literature relevant to the research topics touched by our work.

\textbf{Optical Flow - Energy Minimization models.}
For a long time, optical flow has been cast as a continuous optimization problem through variational frameworks \cite{horn1981determining,black1993framework,zach2007duality}. These approaches involve a data term coupled with regularization terms, and improvements to the former \cite{brox2009large,weinzaepfel2013deepflow} or the latter \cite{ranftl2014non} represented the primary strategy to increase optical flow accuracy for years \cite{sun2010secrets}.
More recent strategies consider optical flow as a discrete optimization problem, despite managing the sizeable 2D search space required to determine corresponding pixels between images \cite{menze2015discrete,chen2016full,xu2017accurate} is challenging.
Until a few years ago \cite{dosovitskiy2015flownet}, early attempts to improve optical flow with deep networks mainly focused on learning more robust data terms by training CNNs to match patches across images  \cite{weinzaepfel2013deepflow,bai2016exploiting,xu2017accurate}.

\textbf{End-to-end Optical Flow.} 
FlowNet \cite{dosovitskiy2015flownet} is the first end-to-end deep architecture proposed for optical flow. Concurrently, to satisfy the massive amount of training data required, synthetic datasets with dense optical flow \gt{} labels were made available \cite{dosovitskiy2015flownet,mayer2016dispnet}. 
Eventually, other architectures \cite{ilg2017flownet2,sun2018pwc,sun2019models,hui18liteflownet,hui20liteflownet2,hui20liteflownet3,teed2020raft} further improved accuracy on popular synthetic \cite{butler2012sintel,mayer2016dispnet} and real \cite{menze2015kitti, geiger2012kitti} benchmarks, with RAFT \cite{teed2020raft} representing state-of-the-art. 

For most existing networks, generalization remains a cause of concerns, in particular when moving from synthetic \cite{dosovitskiy2015flownet,mayer2016dispnet} to real images \cite{geiger2012kitti,menze2015kitti}. With our work, we show how to generate plausible training samples from real, unrelated images allowing for superior generalization.

\textbf{Self-supervised Optical Flow.}
Being \gt{}s hard to obtain for real data, self-supervised strategies allow to relax this requirement \cite{jason2016back,ren2017unsupervised,meister2018unflow}. More recent advances introduced teaching-student frameworks \cite{liu2019ddflow}, occlusion generation \cite{liu2019selflow} and transformed data from augmentation \cite{liu2020learning}. Jonschkowski \etal{} \cite{jonschkowski2020uflow} highlighted the key components to achieve state-of-the-art results in this setting.

Most of these approaches train on unlabeled videos (\eg, from the KITTI 2015 multiview dataset \cite{menze2015kitti}) from the same domain where the evaluation is carried out (\eg, the KITTI 2015 optical flow benchmark).
In contrast, in our work, we relax both constraints of having i) organized video collections and ii) taken in similar domains, achieving superior generalization compared to self-supervised networks.

\textbf{Single Image Depth Estimation.}
In parallel to supervised approaches \cite{Xu_CVPR_2016,Laina_3DV_2016,Fu_2018_CVPR}, many works focused on self-supervised strategies, aimed at replacing \gt{} labels with collections of images, either relying on stereo pairs \cite{godard2017monodepth, tosi2019monoresmatch,watson2019depthhints} or monocular videos \cite{zhou2017unsupervised,godard2019monodepth2,packnet,tosi2020distilled}.
To improve generalization, recent works \cite{li2018megadepth,ranftl2020midas} exploited supervision from a large variety of images and auxiliary strategies such as Multi-View Stereo methods \cite{schoenberger2016sfm}.

Shared by all these methods is the assumption of static scenes, required for reprojection across multiple views. In this paper, we show how a network trained according to such a strategy allows for generating, from still images, training data that well model motions, to train optical flow networks that are effective in presence of moving objects.

\begin{figure}
    \centering
    \includegraphics[trim=1cm 2cm 16cm 0cm,clip,width=0.35\textwidth]{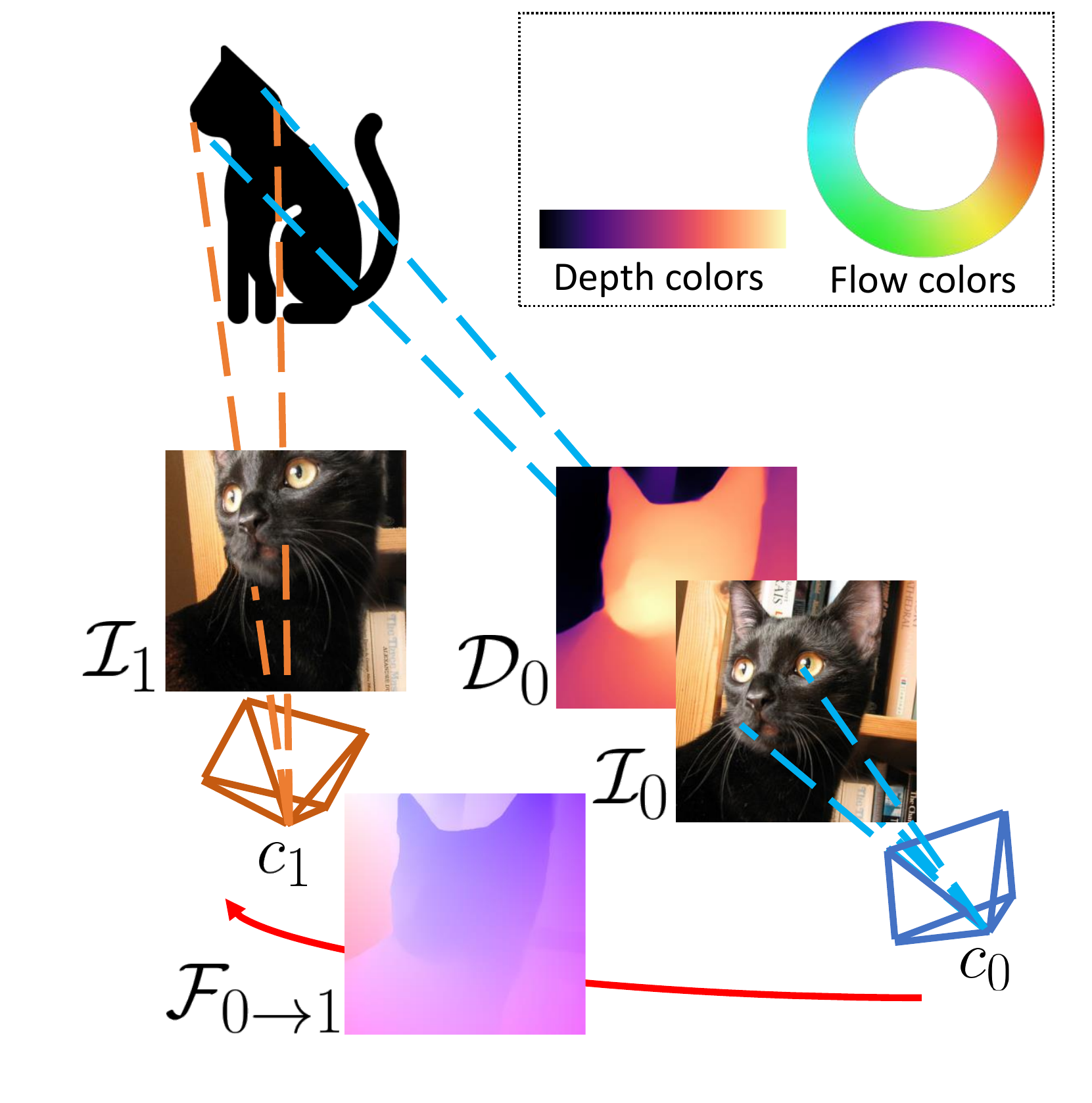}
    \caption{\textbf{Overview of the proposed depthstillation pipeline.} Given a single image $\mathcal{I}_0$ and its estimated depth map $\mathcal{D}_0$, we place the camera in $c_0$ and \textit{virtually} move it (red arrow) towards a new viewpoint $c_1$. From the depth and virtual ego-motion, we obtain optical flow labels $\mathcal{F}_{0\rightarrow 1}$ and a novel $\mathcal{I}_1$ through forward warping.}
    \label{fig:overview}
\end{figure}

\textbf{Novel View Synthesis.}
View synthesis aims at creating new images observed from arbitrary viewpoints starting from a given scene. It is gaining an ever increasing interest in computer vision \cite{yoon2020novel,mildenhall2020nerf,flynn2019deepview,tucker2020singleviewsynthesis,riegler2020FVS}, and it is a fundamental step to address many other tasks, such as video interpolation \cite{jiang2018slowmotion,bao2019depthinterpolation,Niklaus_CVPR_2020} or 3D effects \cite{shih20203dphoto,zhou2018stereomagnification,niklaus2019kenburns}.

Conversely, we focus on creating image pairs and corresponding \gt{} pixel displacements rather than visually pleasant videos. While some of the techniques mentioned above rely on pre-trained flow networks \cite{Niklaus_CVPR_2020}, our goal is to generate data to train these latter.

\textbf{Data distillation through depth estimation.}
Strictly related to our work is \cite{watson2020stereo}, estimating depth from single images to synthesize virtual right views and thus obtain stereo pairs, used to train deep stereo networks.

Despite the analogy of using single image depth estimation, we point out that our goal differs from \cite{watson2020stereo} since we aim at modeling arbitrary motions in the scene (\ie, optical flow) rather than a horizontal pixel displacement between synchronized images (\ie, disparity). 
Purposely, we will describe the additional strategies required to attain, from single still images, the best training data for optical flow networks.

\section{Depthstillation pipeline}
In this section we illustrate our proposed framework to generate new virtual views $\mathcal{I}_1$ from single images $\mathcal{I}_0$, with corresponding dense optical flow \gt{} maps $\mathcal{F}_{0\rightarrow1}$. An overview of our pipeline is shown in Figure \ref{fig:overview}.

\textbf{Virtual camera motion engine.}
Given $\mathcal{I}_0$, an off-the-shelf monocular depth network $\Phi$ is used to estimate its depth map $\mathcal{D}_0$
\begin{equation}
    \mathcal{D}_0 = \Phi(\mathcal{I}_0)
\end{equation}
used to project pixels in $\mathcal{I}_{0}$ to 3D space according to some plausible inverse intrinsics matrix $K^{-1}$. In case the network estimates inverse depth, we bring it to the depth domain first.
$\mathcal{D}_0$ usually shows blurred edges \cite{watson2020stereo,shih20203dphoto}, causing flying pixels in the 3D space that can be easily sharpened via edge-preserving filters \cite{ma2013bilateralfilter}.

We now assume the camera used to frame image $\mathcal{I}_0$ to be at 3D location $c_0$ and apply an arbitrary \textit{virtual} motion, moving it towards a new position $c_1$. To this aim, we generate a plausible rotation $R_1$ by sampling a random triplet of Euler angles and a plausible translation $t_1$ by sampling a random 3D vector. Then, we obtain the transformation matrix $T_{0\rightarrow1}=(R_1|t_1)$ corresponding to such roto-translation.
Thus, we can project our 3D points to the image space through $K$ in order to obtain a new image $\mathcal{I}_1$. This allows to obtain, for each pixel $p_{0}$ in $\mathcal{I}_0$, the coordinates $p_{1}$ of its corresponding pixel in $\mathcal{I}_1$ acquired from viewpoint $c_1$
\begin{equation}\label{eq:pix2pix}
    p_1 \sim KT_{0\rightarrow1}\mathcal{D}_0(p_0)K^{-1}p_0
\end{equation}
and flow $\mathcal{F}_{0\rightarrow1}$ is obtained as the difference between $p_1$ and $p_0$.
We point out that $\mathcal{F}_{0\rightarrow1}$ only models the virtual camera ego-motion, \ie no object has moved independently.
Finally, we obtain the new image $\mathcal{I}_1$ through forward warping.

Forward warping suffers from two well-known problems \cite{watson2020stereo},
that are \textit{collisions} (\ie, multiple pixels from $\mathcal{I}_0$ being warped to the same location in $\mathcal{I}_1$) and \textit{holes} (\ie, pixels in $\mathcal{I}_1$ over which no pixel from $\mathcal{I}_0$ is projected).
To handle collisions, we keep track of pixels $p_1$ having multiple projections $p_0$ in a binary collision mask $\mathcal{M}$ (\ie, collisions are labeled as 1, other pixels as 0) and select, for each, the one having minimum depth according to camera in position $c_1$, \ie the closest, to be displayed in $\mathcal{I}_1$. 

\begin{figure}[t]
    \centering
    \renewcommand{\tabcolsep}{1pt}
    \begin{tabular}{ccccc}
        \includegraphics[height=0.09\textwidth]{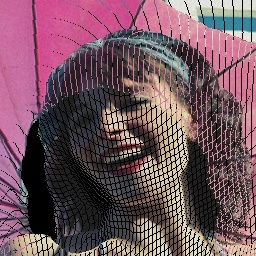} &
        \includegraphics[height=0.09\textwidth]{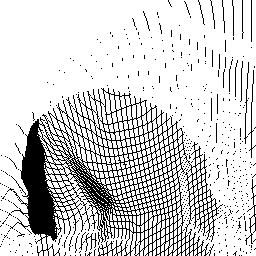} &
        \includegraphics[height=0.09\textwidth]{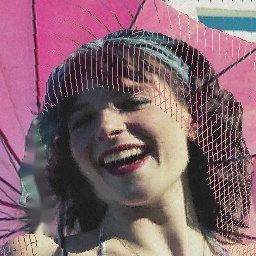} &
        \includegraphics[height=0.09\textwidth]{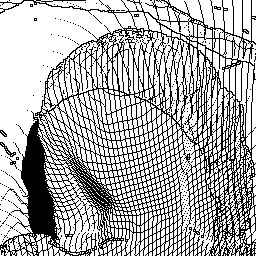} &
        \includegraphics[height=0.09\textwidth]{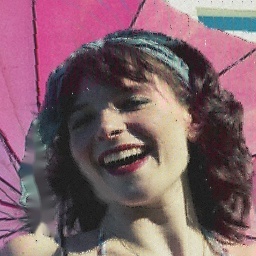} \\
        a) & b) & c) & d) & e) \\
    \end{tabular}
    \caption{\textbf{Hole filling strategies.} From left to right: a) forward-warped image affected by stretching artefacts, b) holes mask $\mathcal{H}$ c) inpainted image, d) collision-augmented holes mask $\mathcal{H'}$ and e) improved inpainted image. Black pixels in $\mathcal{H}$ and $\mathcal{H}'$ are those to be inpainted.}
    \label{fig:holes}
\end{figure}

\textbf{Hole filling.}
Artefacts introduced by holes are more subtle to be solved. Moreover,
applying a 6DoF transformation to the camera plane vastly increases the chance of occurrence of holes compared to the case of 1D camera translations applied to distill stereo pairs \cite{watson2020stereo}.
In particular, in case of larger camera motion/rotations some \textit{stretching} artefacts occur on the foreground objects (and, occasionally, in the background as well) as shown in Figure \ref{fig:holes} a).
To remove these holes, we build a binary hole mask $\mathcal{H}$, as in Figure \ref{fig:holes} b), where we label pixels in $\mathcal{I}_1$ for which no pixel in $\mathcal{I}_0$ is reprojecting on to with 0. Then, a simple inpainting strategy \cite{telea2004inpainting} is usually sufficient to fill them, as reported in Figure \ref{fig:holes} c) on the girl's face. 
Unfortunately, this is not enough in the case of stretching artefacts occurring in a foreground object overlapping a background one. Indeed, in this case, it is very likely that the holes induced by the stretching of the foreground object are filled by pixels in the background. These pixels are not detected by $\mathcal{H}$, causing the bleeding effect shown in Figure \ref{fig:holes} c), where the hair merges with the background umbrella.
Since most of these artefacts occur in non-colliding pixels surrounded by colliding ones, \ie in $\mathcal{M}$ they are labeled as 0 and surrounded by pixels labeled as 1, we can detect them by dilating $\mathcal{M}$ into $\mathcal{M'}$. Then, we define the binary mask $\mathcal{P}$ assigning 1 to pixels having the same label in ($\mathcal{M'}, \mathcal{M}$) and 0 to the remaining (\ie those that become 1 in $\mathcal{M}'$). We finally obtain $\mathcal{H}'$ by multiplying $\mathcal{H}$ and $\mathcal{P}$ 
\begin{equation}
    \mathcal{P} = (\mathcal{M}' == \mathcal{M}), \quad\quad\quad
    \mathcal{H'} = \mathcal{H} \cdot \mathcal{P}
\end{equation}

We can apply the inpainting algorithm to pixels labeled with 0 in $\mathcal{H'}$, shown in Figure \ref{fig:holes} d), to obtain Figure \ref{fig:holes} e), where the foreground-background bleeding does not occur. We report more qualitative examples regarding the different masks in the supplementary material.

We point out how, in large dis-occluded area (\ie, in the proximity of depth boundaries, as shown in Figure \ref{fig:holes} on the left of the person), the inpainting method produces blurred content, as shown in Figure \ref{fig:holes} c) and e). Despite these artefacts, our experiments will prove that hole filling improves the accuracy of trained networks significantly. We report in the supplementary material additional qualitative results concerning the design choices discussed so far.

\textbf{Independent motions.} The pipeline sketched so far models the optical flow field occurring between images acquired in a static environment, \ie consequence of the camera motion, not taking into account possible independently moving objects, very likely to occur in real contexts \cite{menze2015kitti}.
In order to model more realistic simulations, we introduce the possibility of applying different virtual motions to objects extracted from the scene by leveraging an instance segmentation network $\Omega$ for extracting N objects $\Pi_i$, $i \in [1,$N$]$

\begin{equation}
    \Pi = \{\Pi_i, i \in [1,\text{N}]\} = \Omega(\mathcal{I}_0)
\end{equation}
Then, to simulate a motion of the object in the scene, we randomly move the camera from $c_0$ towards a point $c_{\pi_i} \ne c_1$ and its corresponding transformation $T_{0\rightarrow{}{\pi_i}}$ to be applied to object $\Pi_i$. Then, we reproject pixels from $\mathcal{I}_0$ on the image planes of the different cameras. Pixel coordinates in $\mathcal{I}_1$ will be selected according to their belonging to segmented objects or the background as

\begin{equation}
    p_1 \sim \begin{cases} 
        KT_{0\rightarrow1}\mathcal{D}_0(p_0)K^{-1}p_0 & \mbox{if } p_0 \notin \Pi \\ 
        KT_{0\rightarrow{}{\pi_i}}\mathcal{D}_0(p_0)K^{-1}p_0 & \mbox{if } p_0 \in \Pi_i  
    \end{cases}
    \label{eq:instance_final_motion}
\end{equation}
We handle collisions as outlined before, keeping pixels whose depth results lower after motion.
Finally, we obtain optical flow $\mathcal{F}_{0\rightarrow1}$ and image $\mathcal{I}_1$ as aforementioned.

\begin{figure}[t]
    \centering
    \renewcommand{\tabcolsep}{1pt}
    \begin{tabular}{cccc}
        \includegraphics[width=0.10\textwidth]{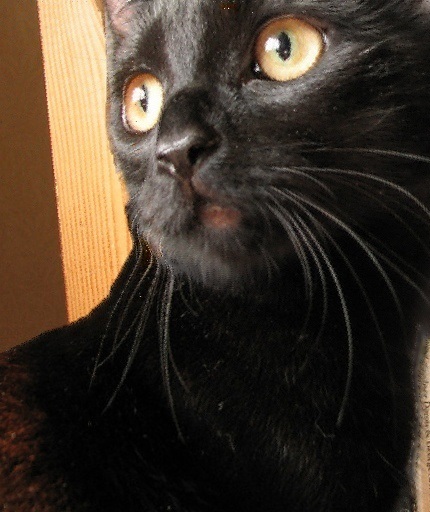} &
        \includegraphics[width=0.10\textwidth]{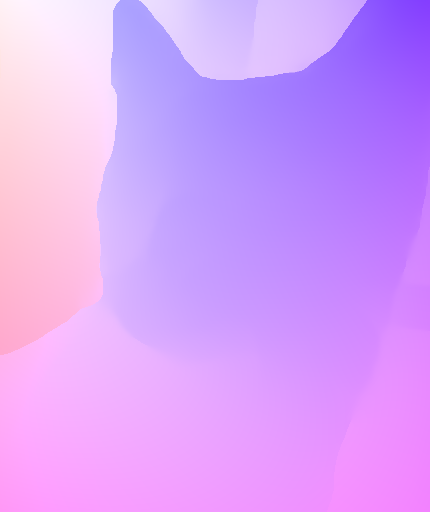} &
        \includegraphics[width=0.10\textwidth]{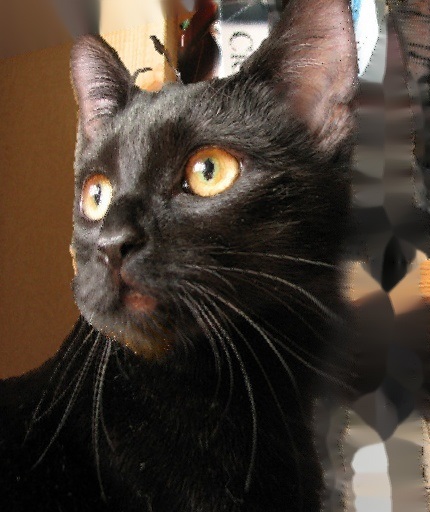} &
        \includegraphics[width=0.10\textwidth]{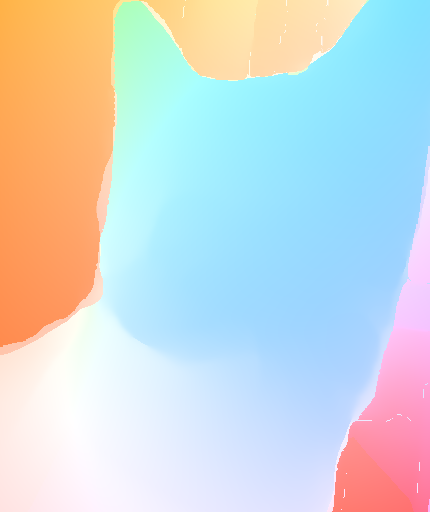} \\
        \includegraphics[width=0.10\textwidth]{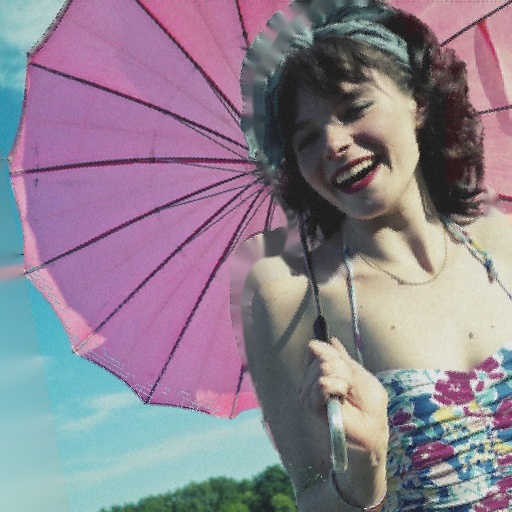} &
        \includegraphics[width=0.10\textwidth]{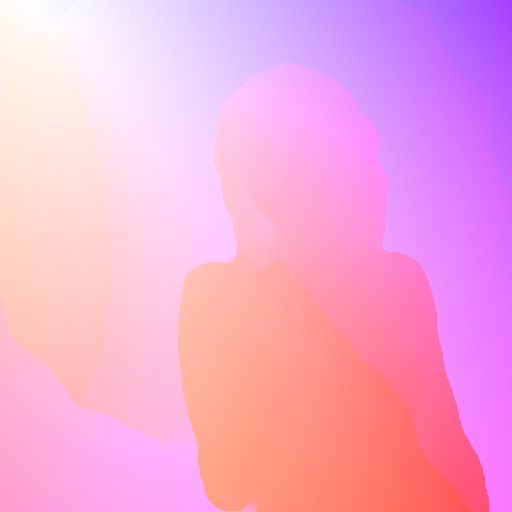} &
        \includegraphics[width=0.10\textwidth]{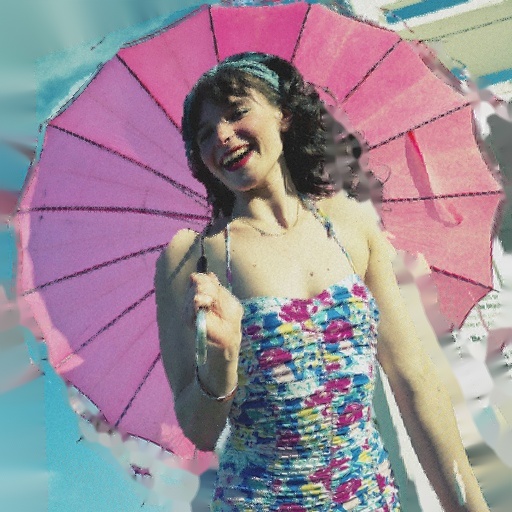} &
        \includegraphics[width=0.10\textwidth]{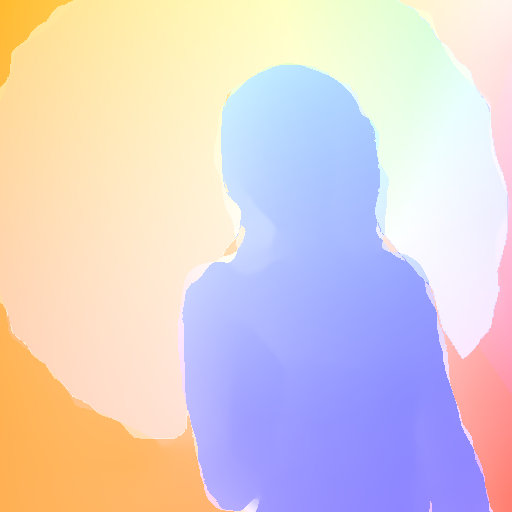} \\        
        a) & b) & c) & d) \\
    \end{tabular}
    \caption{\textbf{Independent motions modeling.} From left to right: a) image generated by only modeling camera motion and b) corresponding optical flow field, c) image generated after segmenting the foreground, which is now subject to a different motion yielding d) a more complex optical flow field.}
    \label{fig:segmentation}
\end{figure}

To be robust to noisy/false detections, \eg in case of tiny blobs accidentally labeled as objects, we rank the objects according to their size, \ie number of pixels, and keep in $\Pi$ only the $n<$N largest objects.
Figure \ref{fig:segmentation} shows two qualitative comparisons between images and flow distilled by merely applying a virtual camera motion, a) and b), and those obtained by segmenting the cat or the person in the foreground and simulating an independent motion, c) and d). Although our formulation simulates moving objects by moving virtual cameras instead, we can notice how the final effect on $\mathcal{I}_1$ and $\mathcal{F}_{0\rightarrow1}$ is equivalent for our purposes.

We point out that, by increasing the number of moving objects, collisions and holes increase. In particular, a higher number of dis-occlusions might appear after applying independent motions, leading to blurry inpainted content, as shown in Figure \ref{fig:segmentation} c) on the top row, on the right of the cat. Besides, shape boundaries may be inconsistent across depth and segmentation predictions, afflicting the truthfulness of the generated image and introducing artefacts (e.g., background pixels moved as part of the foreground). We will see how, although helpful, this approach yields minor improvements compared to the previous two steps performed in our framework, that result crucial for dephtstilling reliable training data. Moreover, segmenting object instances requires an additional network $\Omega$ trained in a supervised manner conversely to single-image depth estimation networks, whereby an extensive literature of self/weakly-supervised approaches exists \cite{godard2017monodepth,godard2019monodepth2,tosi2019monoresmatch,watson2019depthhints}.

\section{Experimental results}

In this section, we describe the experimental setup used to validate our depthstillation pipeline. The source code is available at \url{https://github.com/mattpoggi/depthstillation}.

\subsection{Training datasets}

At first, we describe the datasets used to train the networks considered in our experiments.

\textbf{Chairs (Ch).} FlyingChairs \cite{dosovitskiy2015flownet} is a popular synthetic dataset used to train optical flow models. It contains $22232$ images of chairs moving according to 2D displacement vectors over random backgrounds sampled from Flickr.

\textbf{Things (Th).} The FlyingThings3D dataset \cite{ilg2017flownet2} is a collection of 3D synthetic scenes belonging to the SceneFlow dataset \cite{mayer2016dispnet} and contains a training split made of $19635$ images. Differently from Chairs, objects move in the scene with more complex 3D motions.
State-of-the-art networks usually train in sequence over Chairs and Things (Ch$\rightarrow$Th).

\textbf{COCO dataset.} The COCO dataset \cite{lin2014coco} is a collection of single still images (it provides $\mathcal{I}_0$ only) and \gt{} with labels for tasks such as object detection or panoptic segmentation, but lacks any depth or optical flow annotation. We sample images from the \textit{train2017} split, which contains $118288$ pictures, to generate virtual images and optical flow maps. We dub \textit{dephtstilled} COCO (\textbf{dCOCO}) the training set obtained in such a manner.

\textbf{DAVIS.} The DAVIS dataset \cite{perazzi2016davis} provides high-resolution videos and it is widely used for video object segmentation. Since it does not provide optical flow \gt{} labels, we use all the $10581$ images of the unsupervised 2019 challenge to generate \textbf{dDAVIS} and compare with the state-of-the-art in self-supervised optical flow \cite{jonschkowski2020uflow}.

\subsection{Testing datasets}

We describe here the testing imagery used to evaluate the networks trained on the datasets mentioned above. As metrics, we report the average End-Point Error (EPE) and two error rates, respectively the percentage of pixels with absolute error greater than 3 ($>3$) or both absolute and relative errors greater than 3 and 5\% respectively (Fl), as defined in \cite{menze2015kitti},  on \textit{All} pixels. In every experiment, we will highlight the best results in \textbf{bold} and \underline{underline} the second-best among methods trained in fair conditions.

\textbf{Sintel.} Sintel \cite{butler2012sintel} is a synthetic dataset with \gt{} optical flow maps. We use its training split, counting $1041$ images for both Clean and Final passes, for evaluation.  

\textbf{KITTI.} The KITTI dataset is a popular dataset for autonomous driving with sparse \gt{} values for both depth and optical flow tasks. 
Two versions exist, KITTI 2012 \cite{geiger2012kitti} counting 194 images framing static scenes and KITTI 2015 \cite{menze2015kitti} made of 200 images framing moving objects, in both cases gathered by a car in motion.

\subsection{Implementation Details}\label{implementation_details}

We describe next our pipeline and the networks used for depth estimation and learning optical flow.

\textbf{Depth estimation models.} To obtain dense depth maps from single RGB images, we select two models, respectively MiDaS \cite{ranftl2020midas} and MegaDepth \cite{li2018megadepth}, the former because represents the state-of-the-art for depth estimation in-the-wild and the latter because trained with weaker supervision than MiDaS\footnote{The reader might argue that MiDaS has been trained on labels produced by pre-trained optical flow networks, introducing biases into images generated with our pipeline. However, we point out that optical flow networks are used only to handle negative disparities in stereo images and would not be necessary if, given the minimum negative disparity $d_{min}$, the right image is shifted left by $|d_{min}|$, thus making $d_{min}=0$.}. Next, we will show how the accuracy of networks trained on our data is affected by the depth estimator.

\textbf{Depthstillation pipeline.} To generate virtual images, we convert predicted depths into $[1, 100]$. Given a single image of resolution W${\times}$H, we assume a virtual camera having fixed K, with focals $(f_x,f_y)=0.58$(W,H) and optical center $(c_x,c_y)=0.5$(W,H). To generate $T_{0\rightarrow1}$, we build $t_1$ by sampling three scalars $t_x,t_y,t_z$ in $[-0.2,0.2]$ and $R_1$ by sampling three Euler angles in $[-\frac{\pi}{18},\frac{\pi}{18}]$. To simulate moving objects, we run pre-trained Mask-RCNN \cite{he2017maskrcnn} to select $n=2$ instance masks and generate $t_i$ and $R_i$ sampling respectively in $[-0.1,0.1]$ and $[-\frac{\pi}{36},\frac{\pi}{36}]$ and add them to $R_1$ and $t_1$. Depth maps are sharpened by means of 2 iterations of a $5{\times}5$ bilateral filter, while we dilate $\mathcal{M}$ with a $3{\times}3$ kernel.
We can generate multiple camera motions for any given single image and thus a variety of pairs and \gt{} labels. We will see how playing with the number of images and motions impacts optical flow network accuracy.

\textbf{Optical Flow networks.} To evaluate how effective our distilled images are at training optical flow models, we select two main architectures: RAFT \cite{teed2020raft} and PWC-Net \cite{sun2018pwc}. The first because it represents state-of-the-art architecture for supervised optical flow, already enabling excellent generalization capability. The second because it achieves the best results among self-supervised methods (\eg, UFlow \cite{jonschkowski2020uflow}).
By deploying both architectures, we aim to prove that our method is general and significantly improves generalization in supervised and self-supervised optical flow.
When not otherwise specified, we train RAFT on depthstilled data for 100K steps with a learning rate of 4${\times}10^{-4}$ and weight decay of $10^{-4}$, batch size of $6$ and $496{\times}368$ image crops. This configuration is the largest one fitting into a single NVIDIA Titan X GPU. Following \cite{teed2020raft}, we adopted AdamW as optimizer \cite{loshchilov2017adamw} and applied the same data augmentations and loss functions, while we set $12$ as the number of iterative updates. 
To train PWCNet, we used as optimizer Adam \cite{kingma2014adam}, with an initial learning rate of $1e^{-4}$ and halved after 400K, 600K and 800K steps. We trained our model for 1M steps with a batch size of 8, adopting the multi-scale loss used in \cite{sun2018pwc} for the synthetic pre-training, with the same augmentations and crop size used for RAFT. 

\subsection{Ablation Study}

In this section, we assess the impact of the different components of our pipeline.

\begin{table}[t]
    \centering
    \scalebox{0.62}{
    \begin{tabular}{cccc;RccRccR;RccRcc}
    \hline
    \toprule
         & Depth & Hole & Moving & & \multicolumn{2}{c}{Sintel C.} & & \multicolumn{2}{c}{Sintel F.} & & & \multicolumn{2}{c}{KITTI 12} & & \multicolumn{2}{c}{KITTI 15} \\
        \cline{6-7} \cline{9-10} \cline{13-14} \cline{16-17}
        & est. & fill. & obj. & & EPE & $>3$ & & EPE & $>3$ & & & EPE & Fl & & EPE & Fl  \\
        \toprule
        (A) & \xmark & \xmark & \xmark &  & 5.50 & 18.22 &  & 6.08 & 20.83 & &  & 3.31 & 18.95 & & 10.51 & 35.52 \\
        (B) & \cmark & \xmark & \xmark &  & \underline{2.52} & 7.17 &  & \underline{3.72} & \underline{11.04} & &  & 2.02 & 7.53 & & 4.84 & 16.26 \\
        (C) &\cmark & \cmark & \xmark &  & 2.63 & \underline{7.00} &  & 3.90 & 11.31 & &  & \bfseries1.82 & \underline{6.62} & & \underline{3.81} & \underline{12.42} \\
        (D) & \cmark & \cmark & \cmark &  & \bfseries 2.35 & \bfseries 6.11 &  & \bfseries 3.62 & \bfseries 10.10 &  & & \underline{1.83} & \bfseries6.53 & & \bfseries3.65 & \bfseries 11.98 \\
        \toprule
    \end{tabular}
    }
    \caption{\textbf{Method ablation.} We train RAFT on dCOCO with different configurations of depthstillation: (A) constant depth for each image, (B) adding depth estimated by MiDaS \cite{ranftl2020midas}, (C) adding hole-filling and (D) simulating object motions.}
    \label{tab:ablation}
\end{table}

\begin{table}[t]
    \centering
    \scalebox{0.7}{
    \renewcommand{\tabcolsep}{5pt}
    \begin{tabular}{cc;RccRccR;RccRcc}
    \hline
    \toprule
         & \multirow{1}{*}{Depth Model} & & \multicolumn{2}{c}{Sintel C.} & & \multicolumn{2}{c}{Sintel F.} & & & \multicolumn{2}{c}{KITTI12} & & \multicolumn{2}{c}{KITTI15} \\
         \cline{4-5} \cline{7-8} \cline{11-12} \cline{14-15}
         & & & EPE & $>3$ & & EPE & $>3$ & & & EPE & Fl & & EPE & Fl  \\
        \toprule
        (A) & No depth & & 5.50 & 18.22 &  & 6.08 & 20.83 & &  & 3.31 & 18.95 & & 10.51 & 35.52 \\
        (B) & Megadepth \cite{li2018megadepth} & & \underline{2.91} & \underline{7.51} & & \underline{3.99} & \underline{11.55} & & & \textbf{1.81} & \underline{7.11} & & \underline{4.10} & \underline{13.70} \\
        (C) & MiDaS \cite{ranftl2020midas} & & \bfseries2.63 & \bfseries7.00 &  & \bfseries3.90 & \bfseries11.31 & &  & \underline{1.82} & \bfseries{6.62} & & \bfseries{3.81} & \bfseries{12.42}\\
    \toprule
          
    \end{tabular}
    }
    \caption{\textbf{Impact of depth estimator.} We train RAFT on dCOCO without depth estimation (A), using depth maps provided by MegaDepth (B) or MiDaS (C).}
    \label{tab:depth}
    
\end{table}

\begin{table}[t]
    \centering
    \scalebox{0.7}{
    \renewcommand{\tabcolsep}{2pt}  
    \begin{tabular}{cccc;RccRccR;RccRcc}
    \hline
    \toprule
         & \multicolumn{2}{c}{\# Training samples} & & & \multicolumn{2}{c}{Sintel C.} & & \multicolumn{2}{c}{Sintel F.} & & & \multicolumn{2}{c}{KITTI12} & & \multicolumn{2}{c}{KITTI15} \\
        \cline{2-4} \cline{6-7} \cline{9-10} \cline{13-14} \cline{16-17}
        & Images & Motions & Total & & EPE & $>3$ & & EPE & $>3$ & &  & EPE & Fl & & EPE & Fl  \\
        \hline
        (A) & 4K  & ${\times}$1 & 4K &  & 2.73 & 6.96 & & 3.97 & 11.09 & & & 1.86 & 6.81 & & 3.93 & 12.56 \\
        (B) & 4K  & ${\times}$5 & 20K &  & \underline{2.56} & \underline{6.78} & & \underline{3.88} & \underline{10.99} & & & \bfseries 1.77 & \bfseries 6.62 & & 3.93 & 12.57 \\
        \hline
        (C) & 20K  & ${\times}$1 & 20K &  & 2.63 & 7.00 &  & 3.90 & 11.31 & &  & 1.82 & \bfseries 6.62 & & \bfseries 3.81 & \underline{12.42} \\
        (D) & 20K & ${\times}$5 & 100K &  & \bfseries 2.37 & \bfseries 6.69 & & \bfseries 3.64 & \bfseries 10.73 & & & \underline{1.79} & \underline{6.79} & & \underline{3.82} & \bfseries 12.39 \\
        \toprule
          
    \end{tabular}
    }
    \caption{\textbf{Impact of images and virtual motions.} We train several RAFT models by changing the number of input images taken from COCO and the number of motions depthstilled for each one.}
    \label{tab:number_images}
    
\end{table}

\textbf{Depth, hole filling and moving objects.}
We start by ablating our pipeline to measure the impact of i) estimating depth, ii) applying hole filling to generated images and iii) simulating objects moving independently. This study is carried out by generating virtual views from 20K COCO images, applying a single virtual camera motion for each, by training RAFT \cite{teed2020raft} on them and evaluating the final model on Sintel, KITTI 2012 and KITTI 2015.
Table \ref{tab:ablation} collects the outcome of this evaluation. On row (A), we show the performance achieved by generating images without estimating their depth, thus assuming a constant depth value for all pixels in any image. By moving to row (B), for which we use MiDaS \cite{ranftl2020midas} to estimate depth during the depthstillation process, we can notice considerable improvements in all metrics and datasets, with Fl score often more than halved. Nonetheless, generated images are affected by large holes and this does not allow for optimal performance. By enabling hole filling (C), the trained RAFT further improves its accuracy on real datasets.
Finally, in (D), we show results by simulating objects moving independently, that further improves the results on Sintel.  
The benefit of this latter strategy is consistent on most metrics, although minor on real datasets such as KITTI 2012 and 2015 compared to the improvements obtained by (B) and (C), proving that the simple camera motion combined with depth is enough to obtain a robust optical flow network capable of generalizing to real environments. Moreover, as already pointed out, (D) also requires a trained instance segmentation network, which is hard to obtain for any possible dataset and would consequently constrain our pipeline. Thus, since our primary focus is on real environments, we choose (C) as the configuration for the following experiments.

\textbf{Depth estimation network.} We measure the impact of the depth estimator on our overall data generation pipeline. To this aim, we follow the same protocol of the previous experiments, replacing MiDaS with MegaDepth \cite{li2018megadepth} during the depth estimation step. Table \ref{tab:depth} shows the results of this experiment. We can notice how images generated through MegaDepth (B) allow for training a RAFT model that places in between the one trained on images generated without depth (A) and using MiDaS (C), being much closer to the latter than to the former. This proves that depth is a crucial cue in our pipeline and the accuracy of the optical flow network, as we might expect, increases with the quality of the estimated depth maps, although with minor gains. 

\begin{figure*}[t]
    \centering
    \renewcommand{\tabcolsep}{1pt}
    \begin{tabular}{ccccc}
        \includegraphics[width=0.19\textwidth]{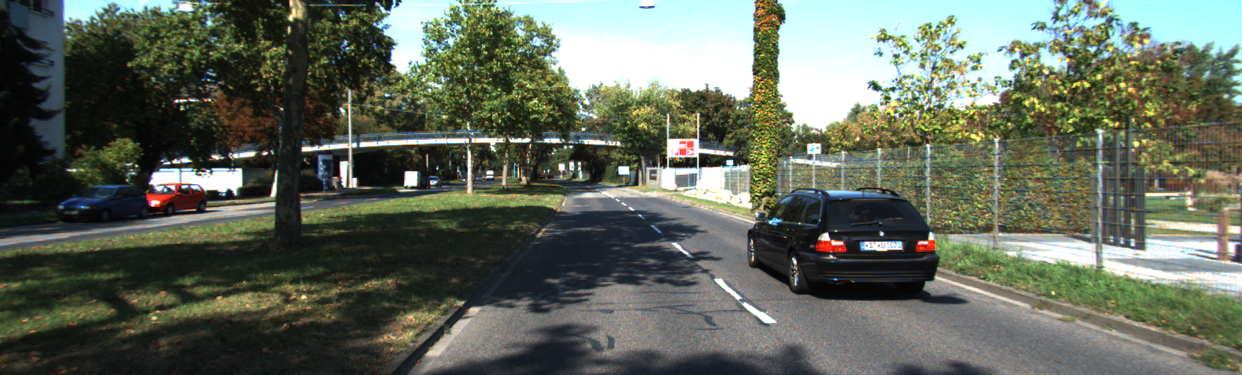} &
        \includegraphics[width=0.19\textwidth]{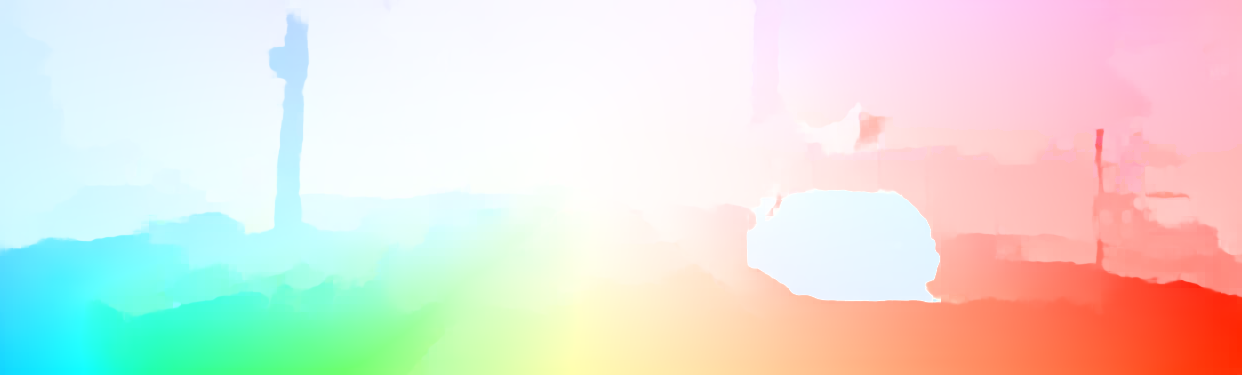} &
        \includegraphics[width=0.19\textwidth]{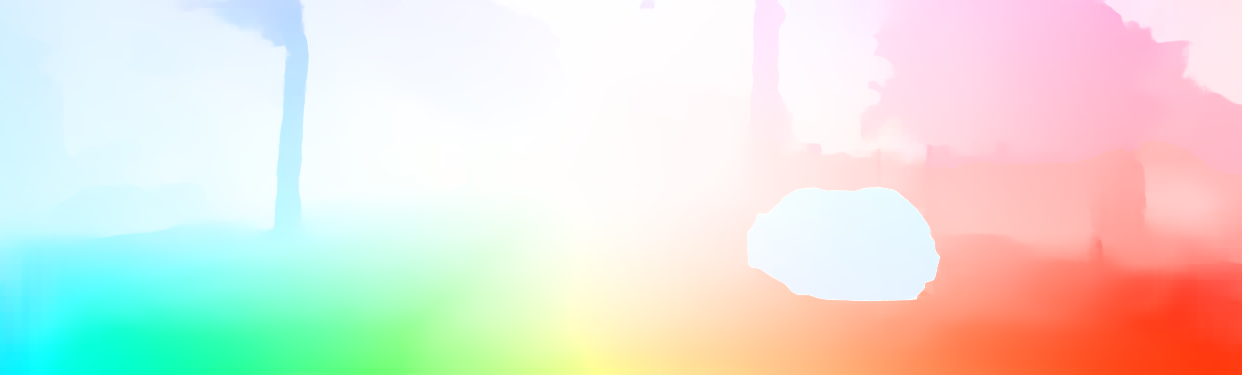} &
        \includegraphics[width=0.19\textwidth]{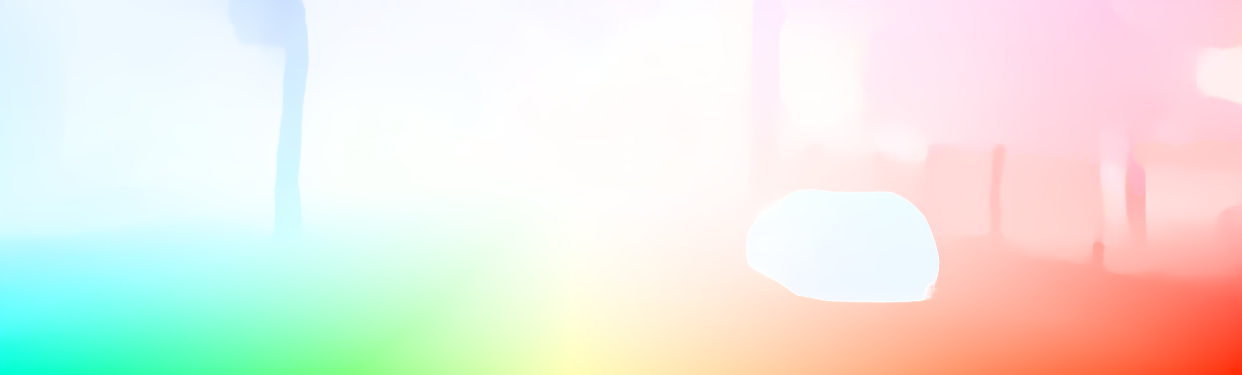} &
        \includegraphics[width=0.19\textwidth]{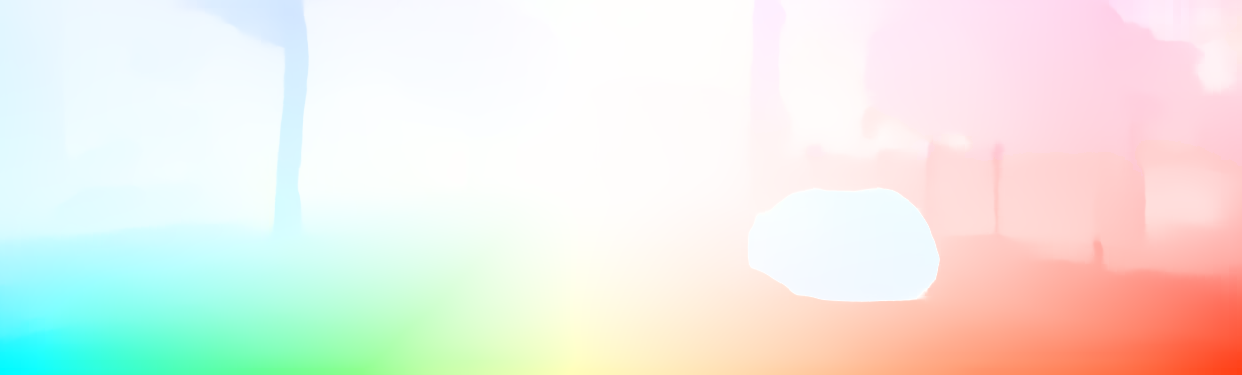} \\
        \includegraphics[width=0.19\textwidth]{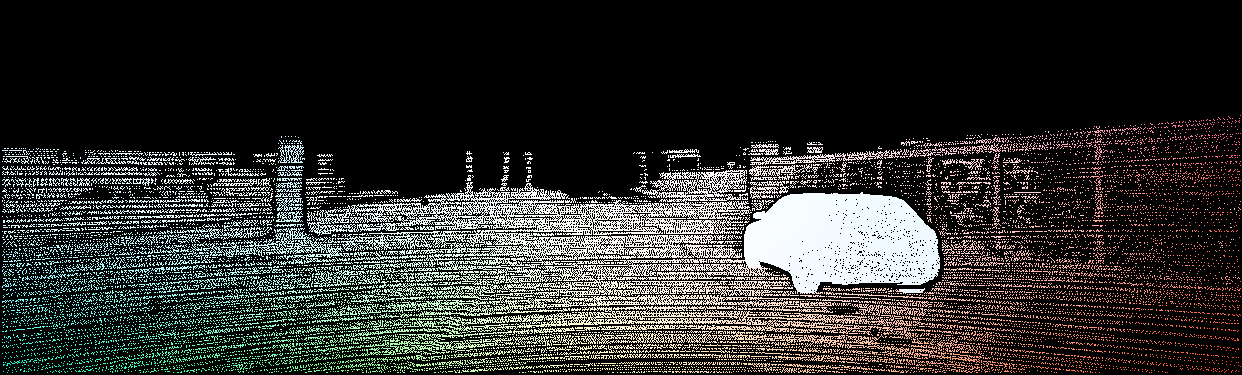} &
        \begin{overpic}[width=0.19\textwidth]{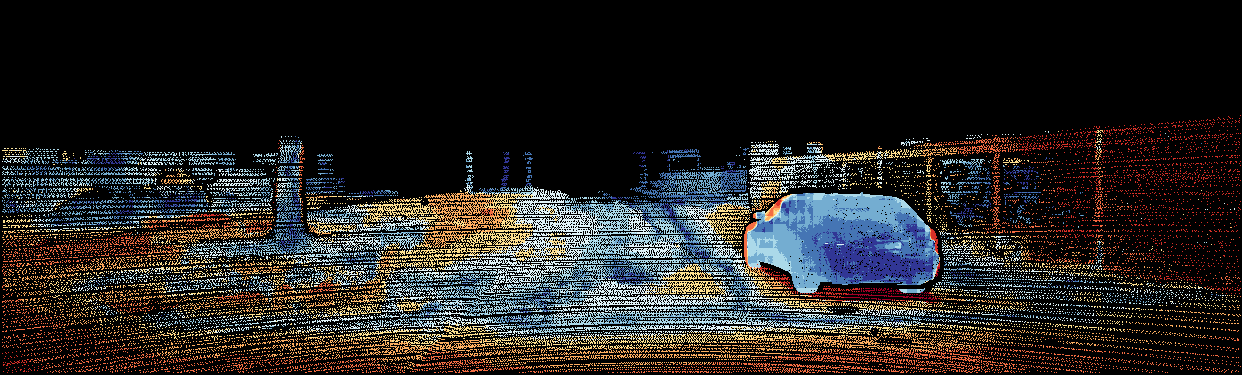}
        \put (4,24) {$\scriptsize\displaystyle\textcolor{white}{\textbf{EPE: 6.43}}$}
        \put (64,24) {$\scriptsize\displaystyle\textcolor{white}{\textbf{Fl: 40.22\%}}$}
        \end{overpic} &
        \begin{overpic}[width=0.19\textwidth]{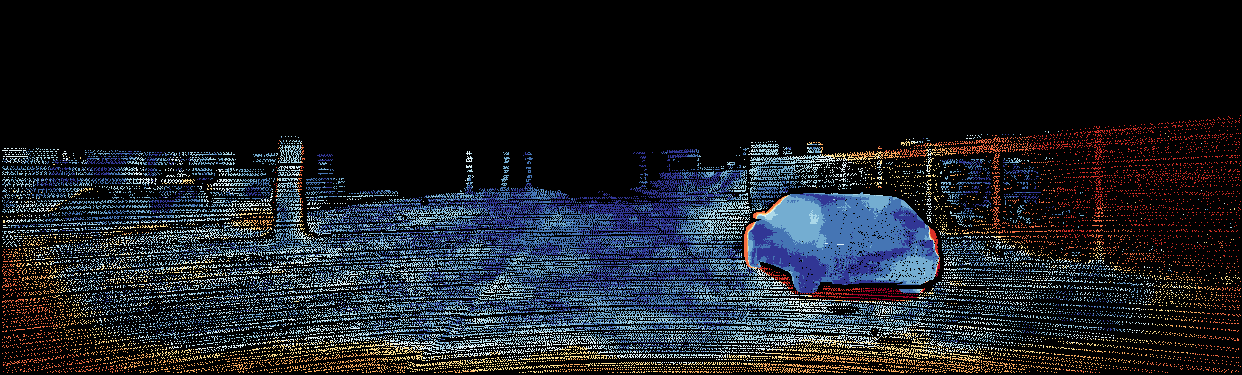}
        \put (4,24) {$\scriptsize\displaystyle\textcolor{white}{\textbf{EPE: 3.91}}$}
        \put (64,24) {$\scriptsize\displaystyle\textcolor{white}{\textbf{Fl: 17.45\%}}$}
        \end{overpic} &
        \begin{overpic}[width=0.19\textwidth]{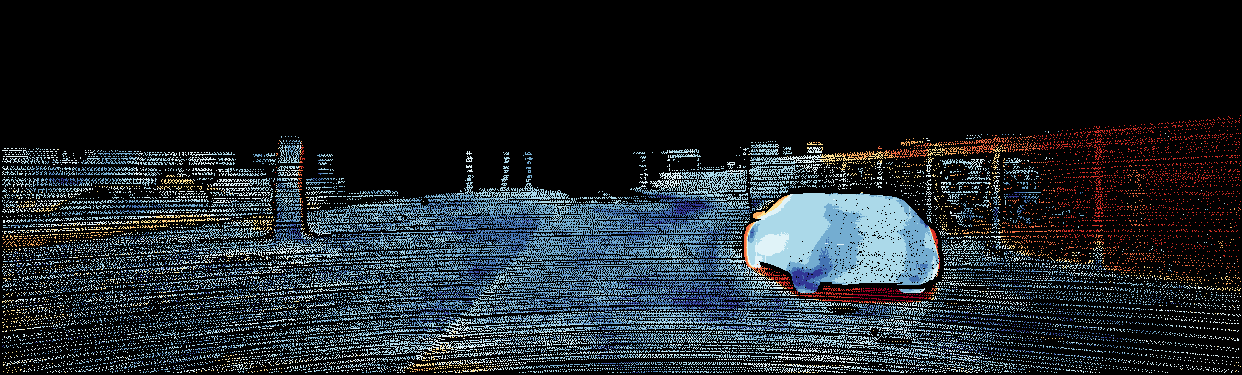}
        \put (4,24) {$\scriptsize\displaystyle\textcolor{white}{\textbf{EPE: 3.04}}$}
        \put (64,24) {$\scriptsize\displaystyle\textcolor{white}{\textbf{Fl: 8.81\%}}$}
        \end{overpic} &
        \begin{overpic}[width=0.19\textwidth]{images/qualitatives/kitti/000019_coco_er.png}
        \put (4,24) {$\scriptsize\displaystyle\textcolor{white}{\textbf{EPE: 3.02}}$}
        \put (64,24) {$\scriptsize\displaystyle\textcolor{white}{\textbf{Fl: 10.23\%}}$}
        \end{overpic} \\
        \includegraphics[width=0.19\textwidth]{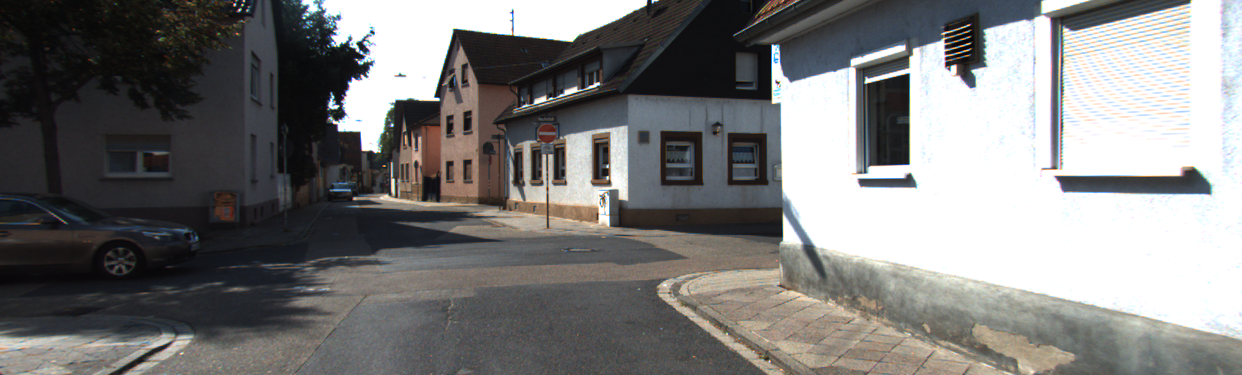} &
        \includegraphics[width=0.19\textwidth]{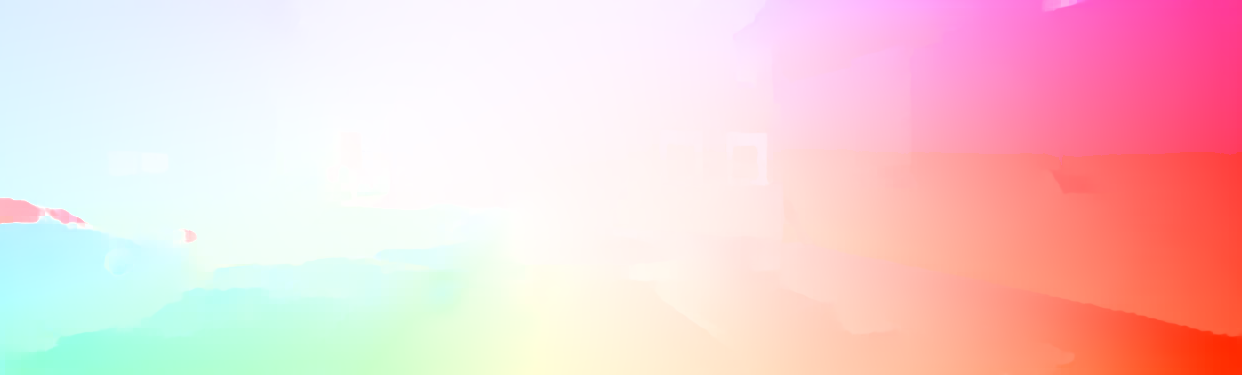} &
        \includegraphics[width=0.19\textwidth]{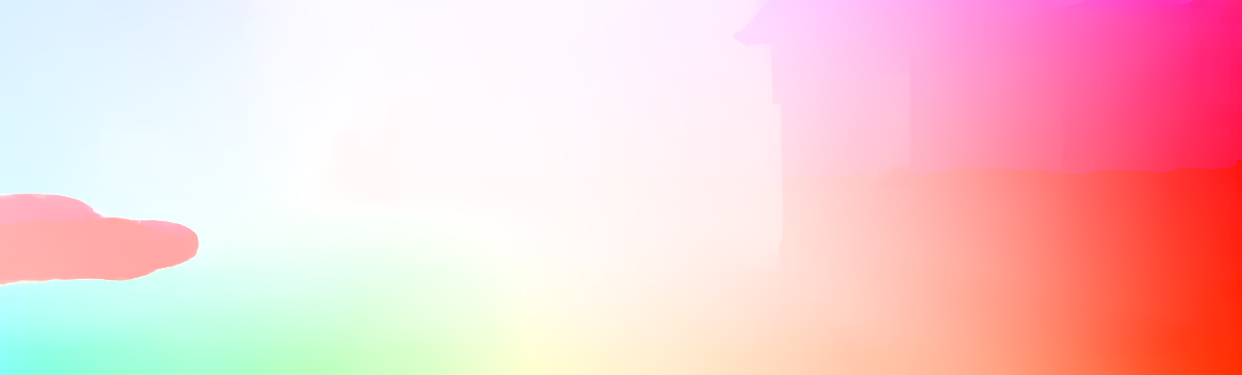} &
        \includegraphics[width=0.19\textwidth]{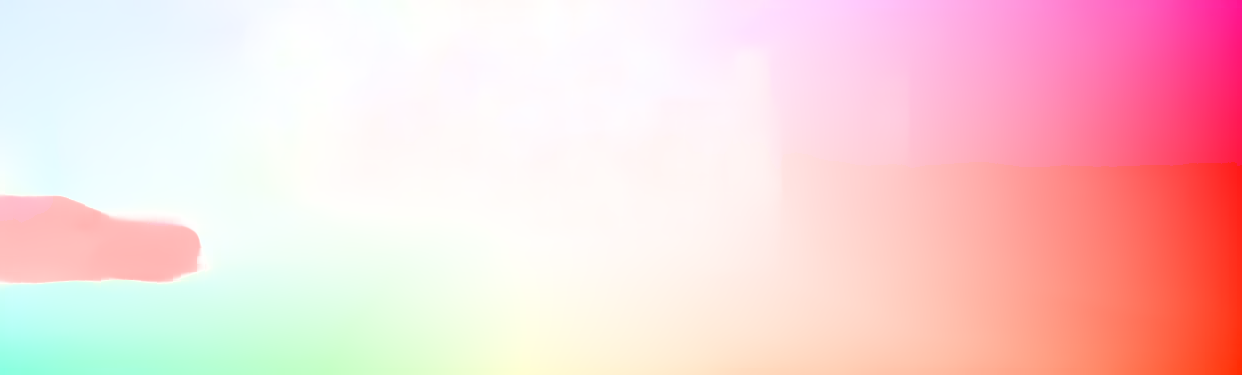} &
        \includegraphics[width=0.19\textwidth]{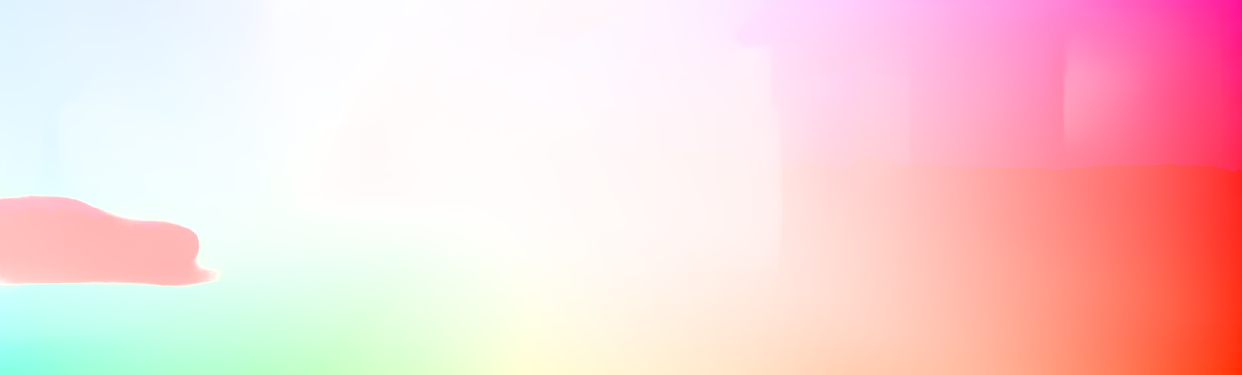} \\
        \includegraphics[width=0.19\textwidth]{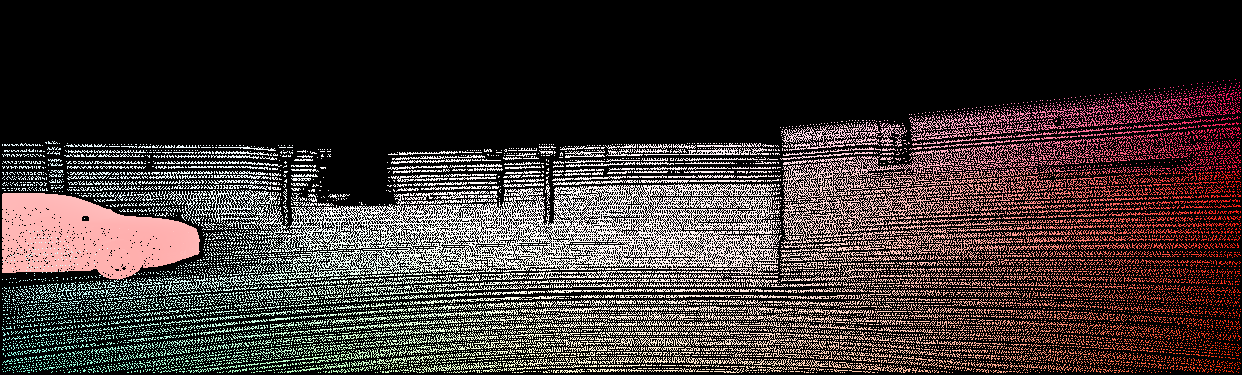} &
        \begin{overpic}[width=0.19\textwidth]{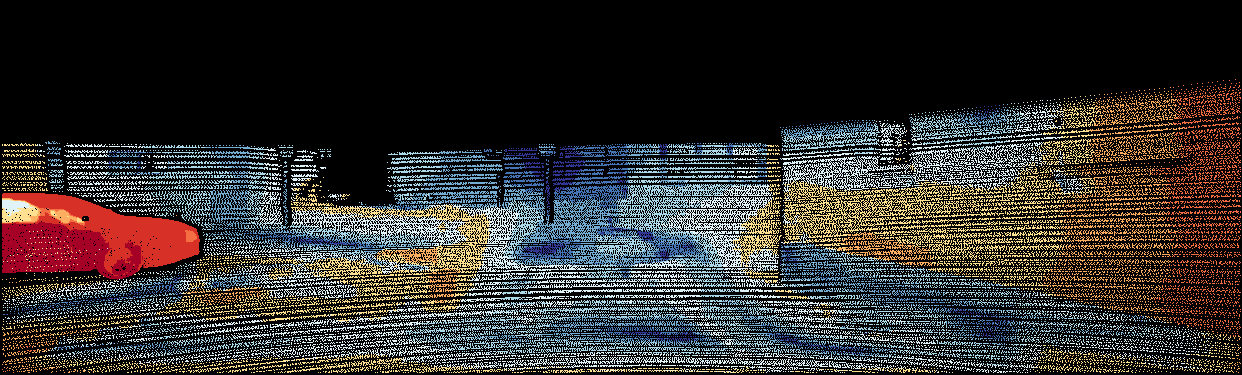}
        \put (4,24) {$\scriptsize\displaystyle\textcolor{white}{\textbf{EPE: 7.21}}$}
        \put (64,24) {$\scriptsize\displaystyle\textcolor{white}{\textbf{Fl: 39.26\%}}$}
        \end{overpic} &
        \begin{overpic}[width=0.19\textwidth]{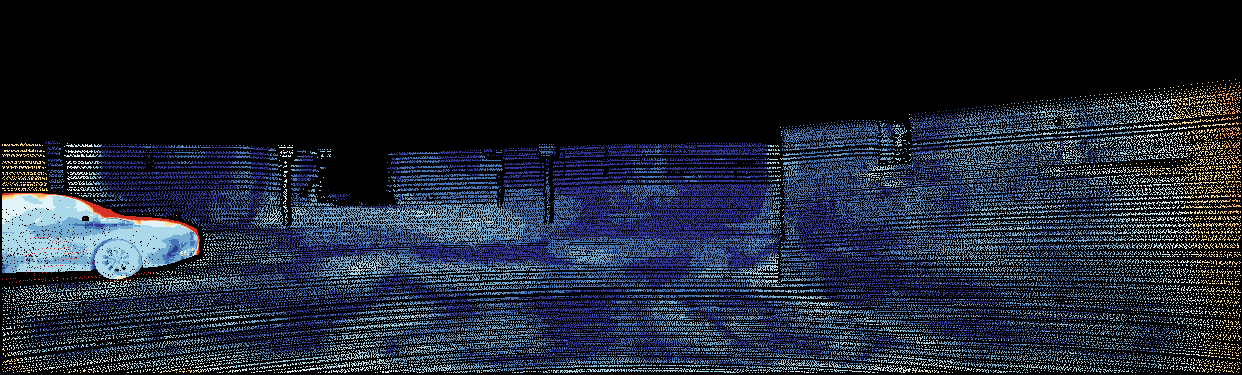}
        \put (4,24) {$\scriptsize\displaystyle\textcolor{white}{\textbf{EPE: 0.95}}$}
        \put (64,24) {$\scriptsize\displaystyle\textcolor{white}{\textbf{Fl: 4.56\%}}$}
        \end{overpic} &
        \begin{overpic}[width=0.19\textwidth]{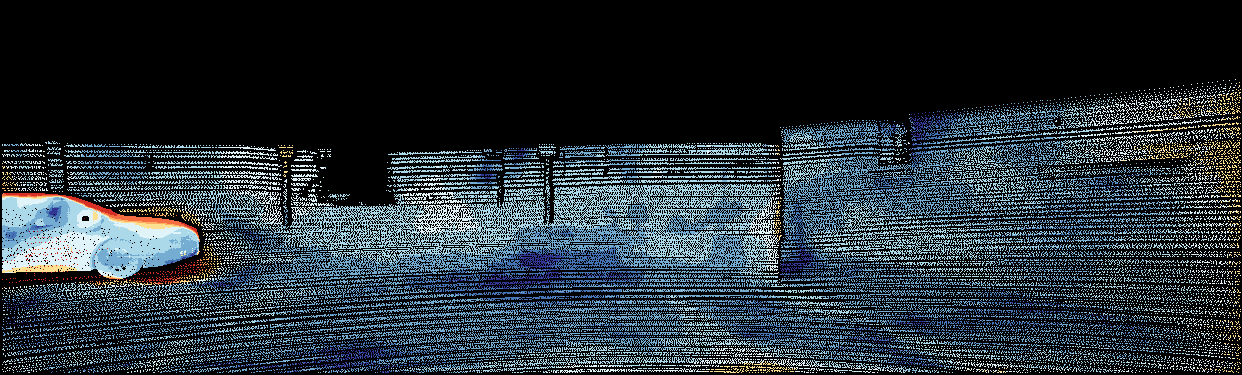}
        \put (4,24) {$\scriptsize\displaystyle\textcolor{white}{\textbf{EPE: 1.32}}$}
        \put (64,24) {$\scriptsize\displaystyle\textcolor{white}{\textbf{Fl: 3.88\%}}$}
        \end{overpic} &
        \begin{overpic}[width=0.19\textwidth]{images/qualitatives/kitti/000085_coco_er.png}
        \put (4,24) {$\scriptsize\displaystyle\textcolor{white}{\textbf{EPE: 1.28}}$}
        \put (64,24) {$\scriptsize\displaystyle\textcolor{white}{\textbf{Fl: 3.91\%}}$}
        \end{overpic} \\        
        a) & b) & c) & d) & e) \\
    \end{tabular}
    
    \caption{\textbf{Qualitative results on the KITTI 2015 training set.} On two rows: a) reference frame (top) and \gt{} flow (bottom), optical flow maps (top) by RAFT trained on b) Ch, c) Ch$\rightarrow$Th, d) dCOCO and e) Ch$\rightarrow$Th$\rightarrow$dCOCO and error maps (bottom).}
    \label{fig:kitti}
\end{figure*}

\textbf{Amount of generated images.} We can increase the amount of data we generate acting on two orthogonal dimensions: the number of images $\mathcal{I}_0$ and the number of virtual motions we simulate for each. Table \ref{tab:number_images} collects the results achieved by several RAFT models trained on a different number of images, obtained by varying the parameters mentioned above. By assuming 4K input images, we can notice how applying 5 virtual motions to each (B) allows a consistent boost on Sintel and KITTI 2012 compared to simulating a single motion each (A), while not improving on KITTI 2015. Interestingly, 4K images already allow for strong generalization to real domains, outperforming the results achieved using synthetic datasets shown in detail in the next section. 
On the other hand, increasing the input images by the same factor ${\times} 5$, yet simulating a single motion (C) leads worse results on Sintel while achieving some improvement on KITTI compared to (A) and (B). This fact highlights that a more variegate image content in the training dataset may be beneficial only for generalization to real environments.
By depthstilling 5 motions, for a total of 100K training samples (D), yields further improvements on Sintel, again with minor impact on KITTI. 
To carry out a fair comparison with synthetic datasets, counting about 20K images each, we will use 20K images and a single virtual motion to depthstill our training data from now on. 

\subsection{Comparison with synthetic datasets}

In this section, we evaluate the effectiveness of our \colorbox{LightYellow}{depthstilled data} versus synthetic datasets \cite{dosovitskiy2015flownet,mayer2016dispnet}.

\textbf{Generalization to real environments.} We start by evaluating the robustness of a network trained on our data when deployed on real datasets. Table \ref{tab:synthetic} shows the performance achieved by RAFT when trained on Chairs (A) and fine-tuned on Things (B) with crop size and settings described in \cite{teed2020raft} to fit in a single GPU, compared to a variant trained on dCOCO, a split of 20K image pairs depthstilled from COCO (C). For completeness, we also report the performance of RAFT models provided by the authors (A\textdagger) and (B\textdagger), trained on 2${\times}$ GPUs and thus not directly comparable with our setting.
We can notice how training on dCOCO (C) allows for much higher generalization on real datasets such as KITTI 2012 and 2015, at the cost of worse performance on the Sintel synthetic dataset. This latter result is not surprising because the images in Things are generated through computer graphics as those in Sintel, while generating virtual images from a real dataset (COCO) leads to superior generalization on real datasets (KITTI 2012 and 2015), also outperforming (A\textdagger) and (B\textdagger) despite the single GPU.

We also train RAFT sequentially on Chairs, Things and dCOCO (D). This setting improves the EPE achieved by (C) on KITTI 2012 and 2015 and turns out much more effective on Sintel with both metrics. This fact suggests that a combination of synthetic images with perfect \gt{} and virtual images with depthstilled labels might be beneficial for generalization purposes.
Figure \ref{fig:kitti} shows some qualitative optical flow predictions and corresponding error maps obtained from the RAFT variants considered in Table \ref{tab:synthetic}. We report additional examples in the supplementary material.

\begin{table}[t]
    \centering
    \renewcommand{\tabcolsep}{2pt}
    \scalebox{0.7}{
    \begin{tabular}{cc;RccRccR;RccRcc}
    \hline
    \toprule
         & Dataset & & \multicolumn{2}{c}{Sintel C.} & & \multicolumn{2}{c}{Sintel F.} & & & \multicolumn{2}{c}{KITTI 12} & & \multicolumn{2}{c}{KITTI 15} \\
        \cline{4-5} \cline{7-8} \cline{11-12} \cline{14-15}
        & & & EPE & $>3$ & & EPE & $>3$ & & & EPE & Fl & & EPE & Fl  \\
        \toprule
        (A\textdagger) & Ch & & 2.26 & 7.35 &  & 4.51 & 12.36 &  &  & 4.66 & 30.54 & & 9.84 & 37.56 \\
        (B\textdagger) & Ch$\rightarrow$Th & & 1.46 & 4.40 &  & 2.79 & 8.10 & &  & 2.15 & 9.30 & & 5.00 & 17.44 \\ 
        \toprule
        (A) & Ch & & 2.36 & 7.70 &  & 4.39 & 12.04 & &  & 5.14 & 34.64 & & 10.77 & 41.08 \\
        (B) & Ch$\rightarrow$Th & & \bfseries{1.64} & \bfseries{4.71} &  & \bfseries{2.83} & \bfseries{8.67} &  & & 2.40 & 10.49 & & 5.62 & 18.71 \\
        \rowcolor{LightYellow}
        (C) & dCOCO & & 2.63 & 7.00 &  & 3.90 & 11.31 & &  & \underline{1.82} & \bfseries 6.62 & & \underline{3.81} & \bfseries 12.42 \\
        \rowcolor{LightYellow}
        (D) & \cellcolor{LightYellow}Ch$\rightarrow$Th$\rightarrow$dCOCO & & \underline{1.88} & \underline{5.31} & & \underline{3.23} & \underline{9.26} & & & \bfseries 1.78 & \underline{7.00} & & \bfseries{3.42} & \underline{13.08} \\
        \toprule
    \end{tabular}
    }
    \caption{\textbf{Comparison with synthetic datasets -- generalization.} Generalization achieved by RAFT when trained on synthetic data (A),(B), on our dCOCO dataset (C) and a combination of both (D). \textdagger{} are obtained with publicly available weights by \cite{teed2020raft} (2${\times}$ GPUs).}
    \label{tab:synthetic}
\end{table}

\begin{table}[t]
    \centering
    \scalebox{0.8}{
    \renewcommand{\tabcolsep}{5pt}
    \begin{tabular}{ccc;RccRcc}
    \hline
    \toprule
         & \multirow{1}{*}{Pre-training} &
         \multirow{1}{*}{Fine-tuning} & & \multicolumn{2}{c}{KITTI12} & & \multicolumn{2}{c}{KITTI15} \\
         \cline{5-6} \cline{8-9}
         & & & & EPE & Fl & & EPE & Fl \\
        \hline
        (A) & Ch & \xmark & & 5.14 & 34.64  & & 15.56 & 47.29 \\
        & Ch & \cmark & & 1.42 & 4.86 & & 2.40 & 8.49 \\
        \hline
        (B) & Ch$\rightarrow$Th & \xmark & & 2.40 & 10.49 & & 9.04 & 25.53 \\
        & Ch$\rightarrow$Th & \cmark & & \underline{1.36} & \underline{4.67} & & \underline{2.22} & \underline{8.09} \\
        \hline
        \rowcolor{LightYellow}
        (C) & dCOCO & \xmark & & 1.82 & 6.62 & & 5.09 & 16.72 \\
        \rowcolor{LightYellow}
        & dCOCO & \cmark & & 1.37 & 4.70 & & 2.76 & 9.15 \\
        \hline
        \rowcolor{LightYellow}
        (D) & Ch$\rightarrow$Th$\rightarrow$dCOCO & \xmark & & 1.78 & 7.00 & & 4.82 & 18.03 \\
        \rowcolor{LightYellow}
        & Ch$\rightarrow$Th$\rightarrow$dCOCO & \cmark & & \bfseries 1.32 & \bfseries 4.54 & & \bfseries 2.21 & \bfseries 7.93 \\

    \toprule
    \end{tabular}
    }
    \caption{\textbf{Comparison with synthetic datasets -- fine-tuning.} Performance of RAFT variants pre-trained on synthetic datasets (A) and (B), on dCOCO (C) or both (D) when fine-tuned on a subset of 160 images from KITTI 2015, tested on KITTI 2012 and the remaining 40 images from KITTI 2015.}
    \label{tab:finetuning}
    
\end{table}

\textbf{Fine-tuning on real data.} We evaluate the effect of pre-training on synthetic images or our generated frames when fine-tuning on a few real data with accurate \gt{}. To this aim, we fine-tune RAFT variants on the first 160 images of the KITTI 2015 training set and evaluate on the remaining 40 and KITTI 2012. We train with a learning rate of $10^{-4}$ and weight decay of $10^{-5}$, batch size of $3$ and $960{\times}288$ image crops, converging after 20K iterations. Table \ref{tab:finetuning} collects the outcome of this experiment. We can notice how variants (A) and (B) trained on synthetic data are greatly improved by the fine-tuning, while (C) achieves slightly lower accuracy after fine-tuning. Despite allowing for much higher generalization to real images, the supervision allowed by our method is \textit{weaker} than the one obtained through real image pairs and perfect \gt{}. Thus, it is not surprising that networks trained from scratch to the end on perfect \gt{} might yield better accuracy. Nonetheless, combining synthetic data with our depthstilled images (D) allows for the best performance, confirming the findings from our previous experiments that a combination of the two worlds -- synthetic data with perfect labels and realistic yet imperfect images and labels -- is beneficial.

\begin{table}[t]
    \centering
    \scalebox{0.75}{
    \renewcommand{\tabcolsep}{1pt}  
    \begin{tabular}{ccc;RccRccR;RccRcc}
    \hline
    \toprule
        & \multirow{1}{*}{Model} & \multirow{1}{*}{Dataset} & &  \multicolumn{2}{c}{Sintel C.} & & \multicolumn{2}{c}{Sintel F.} & & &  \multicolumn{2}{c}{KITTI12} & & \multicolumn{2}{c}{KITTI15} \\
         \cline{5-6} \cline{8-9} \cline{12-13} \cline{15-16}
        &  & & & EPE & $>3$ & & EPE & $>3$ & & & EPE & Fl & & EPE & Fl  \\
        \hline
        (A) & PWCNet & Ch & & 3.33 & - &  & 4.59 & - & & & 5.14 & 28.67 & & 13.20 & 41.79\\
        (B) & PWCNet  & Ch$\rightarrow$Th & & \bfseries 2.55 & - &  & \bfseries 3.93 & - & & & 4.14 & 21.38 & & 10.35 & 33.67\\
        \rowcolor{LightYellow}
        (C) & PWCNet  & dCOCO & & 4.14 & 11.54 &  & 5.57 & 15.58 & & & \bfseries 3.16 & \bfseries 13.30 & & \bfseries 8.49 & \bfseries 26.06 \\
        \toprule
        \rowcolor{LightYellow}
        (D) & RAFT  & dCOCO & & 2.63 & 7.00 &  & 3.90 & 11.31 & &  & 1.82 & 6.62 & & 3.81 & 12.42 \\
    \toprule
          
    \end{tabular}
    }
    \caption{\textbf{Impact of depthstillation on different architectures}. Evaluation on PWCNet and RAFT. Entries with "-" are not provided in the original paper.}
    \label{tab:network}
    
\end{table}

\textbf{Impact on different optical flow networks.} To prove that the superior generalization we achieve is enabled by our data rather than a specific architecture such as RAFT, we also train PWCNet \cite{sun2018pwc} on the 20K images generated from COCO. Table \ref{tab:network} shows how PWCNet trained on dCOCO (C) dramatically outperforms the original variants trained on Chairs (A) and fine-tuned on Things (B) when testing on real data, at the cost of lower performance on Sintel synthetic images, substantially confirming our findings from previous experiments with RAFT, reported in the table for comparison (D). This fact proves that our data, generated from single yet realistic still images, significantly improves generalization to real data independently from the optical flow model trained.

\subsection{Comparison with self-supervision from videos}

Given the rich literature about self-supervised optical flow \cite{meister2018unflow,liu2019ddflow,liu2019selflow,jonschkowski2020uflow}, we compare our strategy with state-of-the-art practises for self-supervised optical flow \cite{jonschkowski2020uflow}. 

\textbf{Generalization.}
In contrast to most works in this field that train and test in the same domain \cite{meister2018unflow,liu2019ddflow,liu2019selflow,jonschkowski2020uflow}, we inquire about how well networks trained in a self-supervised manner or leveraging our proposal transfer across different real datasets.
To this aim, we adopt DAVIS \cite{perazzi2016davis} for training and evaluate on KITTI 2012 and 2015 as in the previous experiments. 
To train UFlow \cite{jonschkowski2020uflow}, we use the official code provided by the authors. In particular, we trained the model on the entire DAVIS dataset for 1M steps, using a batch size of 1 as suggested in \cite{jonschkowski2020uflow}, $512{\times}384$ resized images and letting unchanged other configuration parameters in order to replicate the authors' settings. Being UFlow based on PWCNet, we train from scratch another instance of PWCNet on dDAVIS for the same number of steps with a batch of 8 over depthstilled images and labels. The learning rate scheduling is the same highlighted in section \ref{implementation_details}, while the crop is $512{\times}384$. This way, we evaluate how well a PWCNet trained on depthstilled data transfers to other datasets compared to a model trained on real videos framing the same image content of the depthstilled images.
Table \ref{tab:self_supervised} collects the outcome of this experiment. We can notice how the PWCNet model trained on dDAVIS (B) transfers much better to the KITTI 2012 and 2015 datasets compared to UFlow trained on the real DAVIS (A), thanks to the stronger supervision from the distilled optical flow labels. For the sake of completeness, we also report the results achieved by RAFT (C) trained on the same data, confirming to be superior. 

\textbf{Limitations.} Our pipeline has some obvious limitations. Indeed, the training samples we generate are far from being utterly realistic because cannot model some behaviors, such as the large 3D rotation of objects in the scene, frequently found in real videos. Thus, despite the strong generalization we achieve compared to self-supervision, real videos allow for much better specialization when training and testing in the same domain. 
As shown in Table \ref{tab:self_supervised_kitti},  UFlow trained on the 4K images of the KITTI multiview dataset (A) performs much better than PWCNet trained on $960\times288$ crops from dKITTI (B), a set of about 4K images depthstilled from KITTI 2015 multiview testing set. On the other hand, RAFT trained on dKITTI with the same crop size (C) gets closer to UFlow, thanks to the more effective architecture. 

This lower specialization is also due to the completely random motions we depthstill. In contrast, KITTI motions consist of a much smaller subset (\ie mostly forward translations or steerings) dominant in the real KITTI multiview split, yet rarely occurring in dKITTI.

As take-home message, our depthstillation strategy effectively addresses the scarcity of training data, \eg when annotated images or not-annotated videos of the target environment are not available, yielding superior generalization compared to existing practices. Moreover, it is complementary to domain-specific real training data with labels, seldom ever available in practice.

\begin{table}[t]
    \centering
    \scalebox{0.8}{
    \renewcommand{\tabcolsep}{2pt} 
    \begin{tabular}{ccc;RccRcc}
        \toprule
        & \multirow{1}{*}{Model} & \multirow{1}{*}{Dataset} & &  \multicolumn{2}{c}{KITTI12} & & \multicolumn{2}{c}{KITTI15} \\
         \cline{5-6} \cline{8-9} 
        & & & & EPE & Fl & & EPE & Fl  \\
        \hline
        (A) & UFlow & DAVIS & & 3.49 & 14.54 & & 9.52 & 25.52   \\
        \rowcolor{LightYellow}
        (B) & PWCNet & dDAVIS & & \bfseries 2.81 & \bfseries 11.29 & & \bfseries 6.88 & \bfseries 21.87 \\
        \toprule
        \rowcolor{LightYellow}
        (C) & RAFT & dDAVIS & & 1.78 & 6.85 & & 3.80  & 13.22 \\
        \toprule
    \end{tabular}
    }
    \caption{\textbf{Comparison between self-supervision and depthstillation -- generalization.} Effectiveness of the two strategies when evaluated on unseen data (KITTI 2012 and 2015).}
    \label{tab:self_supervised}
\end{table}

\begin{table}[t]
    \centering
    \scalebox{0.8}{
    \renewcommand{\tabcolsep}{2pt} 
    \begin{tabular}{ccc;RccRcc}
        \hline
        \toprule
        & \multirow{1}{*}{Model} & \multirow{1}{*}{Dataset} & &  \multicolumn{2}{c}{KITTI12} & & \multicolumn{2}{c}{KITTI15} \\
         \cline{5-6} \cline{8-9} 
        & & & & EPE & Fl & & EPE & Fl  \\
        \hline
        (A) & UFlow & KITTI & & - & - & & \textbf{3.08} & \textbf{10.00}   \\
        \rowcolor{LightYellow}
        (B) & PWCNet & dKITTI & & 2.64 & 9.43 & & 7.92 & 22.17 \\
        \toprule
        \rowcolor{LightYellow}
        (C) & RAFT & dKITTI & & 1.76 & 5.91 & & 4.01 & 13.35 \\
        \toprule
    \end{tabular}
    }
    \caption{\textbf{Comparison between self-supervision and depthstillation -- specialization.} Effectiveness of the two strategies when training and testing on similar data (KITTI 2015). Entries with "-" are not provided in the original paper.}
    \label{tab:self_supervised_kitti}
\end{table}

\section{Conclusion}
We proposed a new strategy named, \textit{Depthstillation}, to distill dense optical flow \gt{} maps from single still images and create novel virtual views, by leveraging the depth provided by a pre-trained monocular network. 
Through extensive experiments, we showed how it allows for training state-of-the-art optical flow networks \cite{sun2018pwc,teed2020raft}, leading to models that better generalize to real data compared to the use of synthetic images or self-supervision from videos framing different content, while suffering at specialization.
Depthstillation is a powerful solution when domain-specific training data is not available, as occurs in most practical applications in-the-wild.

\textbf{Acknowledgement.} We gratefully acknowledge the support of NVIDIA Corporation with the donation of the Titan Xp GPU used for this research.

{\small
\bibliographystyle{ieee_fullname}
\bibliography{egbib}
}

\end{document}